\documentclass[11pt]{article}
\usepackage[letterpaper,margin=1in]{geometry}
\usepackage{times}
\usepackage[labelfont=bf,font=small]{caption}

\usepackage{amsmath,amsfonts,bm}

\def\eqref#1{equation~\ref{#1}}

\def\1{\bm{1}}

\DeclareMathAlphabet{\mathsfit}{\encodingdefault}{\sfdefault}{m}{sl}
\SetMathAlphabet{\mathsfit}{bold}{\encodingdefault}{\sfdefault}{bx}{n}

\usepackage{natbib}
\usepackage{hyperref}
\usepackage{url}
\usepackage{graphicx}
\usepackage{amsmath,amssymb}
\usepackage{booktabs}
\usepackage{colortbl}
\usepackage{wrapfig}
\usepackage{algorithm}
\usepackage{algpseudocode}
\algrenewcommand\Call[2]{\textproc{#1}\if\relax\detokenize{#2}\relax\else(#2)\fi}
\renewcommand{\algorithmicrequire}{\textbf{Input:}}
\renewcommand{\algorithmicensure}{\textbf{Output:}}
\usepackage{array}
\usepackage{xcolor}
\usepackage{ragged2e}
\newcommand{\method}{FLASH}
\newcommand{\methodLong}{\textbf{F}ull-attention \textbf{L}ow-resolution \textbf{A}nchor with \textbf{S}parse High-resolution \textbf{H}alo attention}
\newcommand{\LN}{L_N}
\newcommand{\LR}{L_R}
\newcommand{\HN}{H_N}
\newcommand{\HR}{H_R}
\newcommand{\parent}{\pi}

\title{Efficient and Faithful High-Resolution\\Image Editing with Sparse Cross-Resolution Denoising}

\author{%
  Maosen Zhao$^{1,2}$ \quad
  Xianfang Zeng$^{3}$ \quad
  Pengtao Chen$^{2}$ \quad
  Peng Ye$^{4}$\\[2pt]
  Gang YU$^{3}$ \quad
  Jiayuan Fan$^{2}$ \quad
  Tao Chen$^{1,2}$\\[6pt]
  $^{1}$Shanghai Innovation Institute\\[2pt]
  $^{2}$Fudan University\\[2pt]
  $^{3}$StepFun\\[2pt]
  $^{4}$The Chinese University of Hong Kong%
}
\date{}

\begin{document}
\maketitle

\begin{abstract}
Instruction-based image editors still primarily operate at 1K, whereas practical editing increasingly requires faithful 2K/4K outputs. Dense high-resolution adaptation preserves the editing trajectory but incurs expensive target--source attention; editing at 1K followed by external refinement is cheaper but can separate detail restoration from the editing trajectory. We introduce \method{}, a native high-resolution editor based on \emph{sparse cross-resolution denoising}. A frozen 1K branch provides a semantic and structural anchor, while a target-resolution branch reconstructs local detail within the same Transformer. An asymmetric topology retains full 1K attention, guides high-resolution queries with spatially aligned low-resolution noisy-target states, and confines high-resolution target--reference interaction to local Halo neighborhoods. Resolution-specific diffusion-state mapping and spatial YaRN adapt the pretrained editor to larger grids. An optional, training-free progressive path replaces early high-resolution denoising with a 1K prefix and completes refinement in the same editor. On VINS-4KEval, the direct model achieves the strongest structural fidelity among the compared methods at both 2K and 4K. A separate 4K forward benchmark with \(R=1\) records \(4.2\times\) faster total forward execution and \(25.3\times\) faster attention than dense target--reference attention. Progressive inference reduces end-to-end 4K latency by \(30.5\%\), with a modest quality cost.
\end{abstract}

\begin{center}
\small \textbf{Github:} \url{https://github.com/FDUPeter/FLASH}
\end{center}

\section{Introduction}

Modern diffusion editors follow open-ended instructions while preserving source content \citep{brooks2023instructpix2pix,liu2025step1x,wu2025qwen}. Practical applications also demand readable text, fine textures, and precise local structure at 2K and 4K. Yet editing models and benchmarks remain concentrated at 1K or below, even as high-resolution generation advances \citep{chen2026vins}. Extending editing beyond 1K therefore requires both target-scale detail and reliable source preservation.

\begin{figure*}[t]
\centering
\includegraphics[width=\columnwidth]{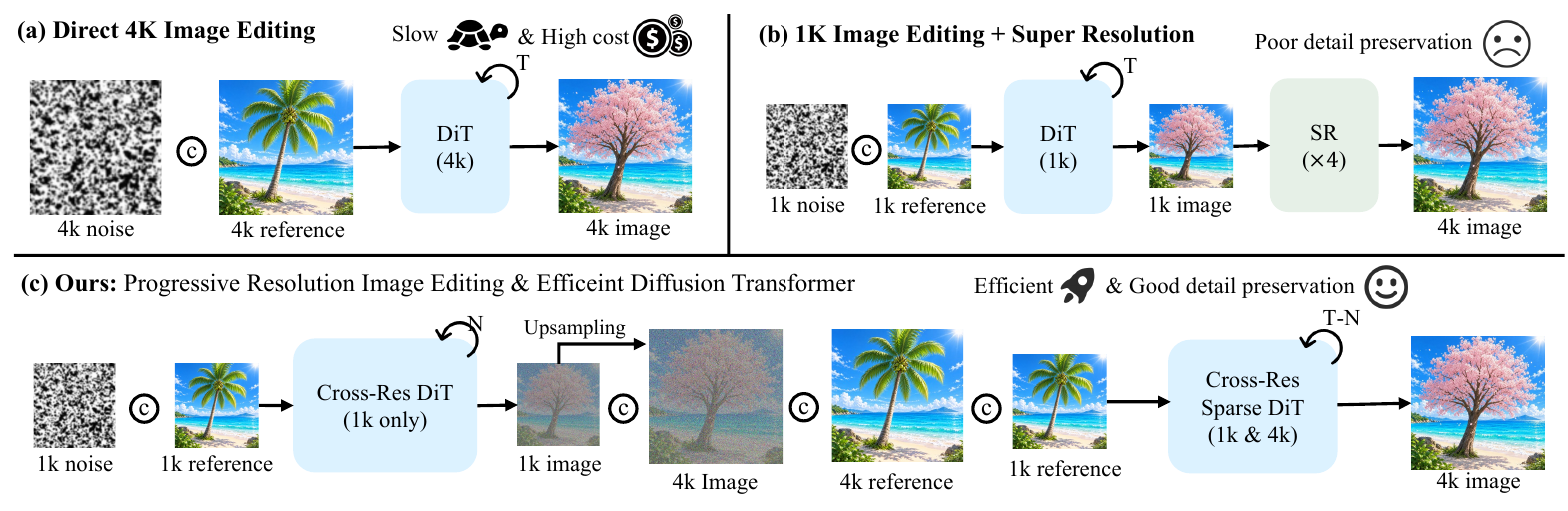}
\caption{Three routes to high-resolution editing: dense native 4K editing,
1K editing with super-resolution, and our \method{}
framework.}
\label{fig:teaser}
\end{figure*}

\begin{wrapfigure}{r}{0.48\columnwidth}
\centering
\includegraphics[width=\linewidth]{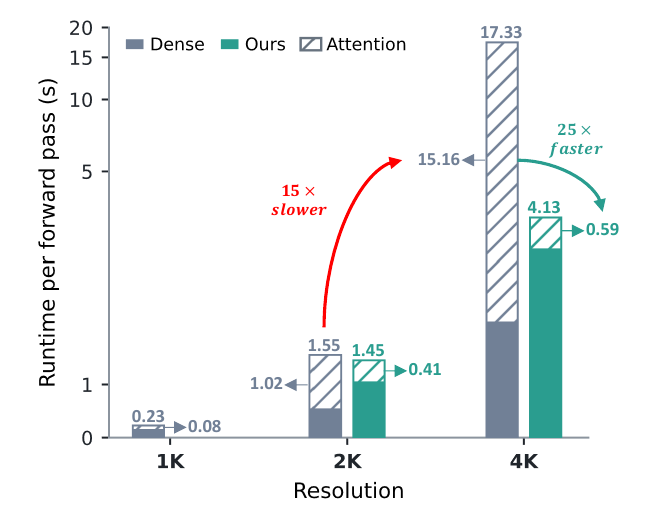}
\caption{Per-forward runtime across resolutions. Each bar shows attention
and other forward computation. Dense attention uses target--reference
sequences; the separate sparse diagnostic uses \(R=1\) and \(h=n\).
At 4K, it is \(4.2\times\) faster overall and \(25.3\times\) faster in
attention. These timings do not represent the main \(R=3\) configuration.}
\label{fig:dense_scaling}
\end{wrapfigure}

Two common approaches trade quality against computation. Dense native adaptation keeps editing and detail synthesis in one denoising trajectory, but its target--source sequences grow rapidly. With fixed latent downsampling and tokenization, scaling both image dimensions by \(k\) increases spatial-token counts by approximately \(k^2\). A dense 4K baseline already takes about 17 seconds per Transformer forward (Fig.~\ref{fig:dense_scaling}), and denoising repeats this cost at every step. Editing at 1K and applying external super-resolution reduces editor cost, but moves detail recovery outside the editing trajectory and can distort source-specific content. We seek a third route: faithful full-canvas 2K/4K editing without dense target-resolution interaction (Fig.~\ref{fig:teaser}).

Our central observation is that the two resolutions serve complementary roles. A capable 1K editor establishes the instruction-conditioned transformation, global layout, and principal edit structure. The high-resolution grid is needed to recover textures, thin boundaries, and structures that cannot be represented at 1K. We therefore use the evolving 1K edit as an anchor for target-resolution detail generation. Both trajectories remain within the same editor: the low-resolution branch forms the global edit, and the high-resolution branch refines it on local support.

\method{} (\methodLong) implements this division through sparse cross-resolution denoising. Its Transformer jointly processes text, low-resolution target--source tokens, and their high-resolution counterparts. A full-attention 1K anchor maintains global editing behavior. Local Halo attention supplies spatially aligned high-resolution source details, while aligned low-to-high guidance transfers the evolving 1K edit to the target grid. Resolution-dependent timestep mapping and spatial RoPE adaptation make both scales compatible with the pretrained editor \citep{peng2024yarn}.

We further reduce temporal redundancy with optional progressive inference. Early high-noise steps run at 1K; the model then lifts a clean estimate through RGB space, re-noises it at the paired high-resolution state, and completes the remaining steps with the same cross-resolution Transformer. High-resolution computation is thus reserved for later refinement. The lifted image is an intermediate state, not the final output of a separate super-resolution stage: subsequent denoising remains conditioned on the instruction, source image, and evolving 1K edit.

Our contributions are threefold:
\begin{enumerate}
    \item We formulate native high-resolution editing as low-resolution-guided denoising, separating global edit formation from target-scale detail generation within one model.
    \item We develop \method{} and a training-free progressive inference path to reduce spatial and temporal redundancy in high-resolution editing.
    \item We evaluate source fidelity at 2K/4K on VINS-4KEval and examine the spatial support and progressive transition. The direct model leads the compared methods in HaarPSI, keep-region mPSNR, and mSSIM at both resolutions. Progressive inference further reduces end-to-end 4K latency by \(30.5\%\).
\end{enumerate}

\section{Related Work}

\subsection{Diffusion-Based Image Editing}

Instruction-based diffusion editing has evolved toward general-purpose source-conditioned generation. InstructPix2Pix established feed-forward instruction following; MagicBrush strengthened human-annotated supervision, and MGIE added multimodal reasoning for underspecified requests \citep{brooks2023instructpix2pix,zhang2023magicbrush,fu2024guiding}. Flow- and DiT-based editors such as Step1X-Edit, Qwen-Image-Edit, and FLUX.1 Kontext further improve instruction fidelity and source consistency \citep{liu2025step1x,wu2025qwen,labs2025flux}. Their training and evaluation nevertheless remain concentrated around 1K \citep{chen2026vins}. Dense target--reference interaction becomes expensive at 2K/4K; our approach retains the 1K editor as an anchor and adapts target-resolution detail modeling.

\subsection{Efficient High-Resolution Generation and Editing}

High-resolution diffusion generation extends pretrained models through receptive-field correction, progressive multiscale synthesis, overlapping diffusion paths, and hierarchical or windowed computation \citep{he2024scalecrafter,du2024demofusion,bar2023multidiffusion,zhang2024hidiffusion,crowson2024scalable}. These approaches establish that low-resolution priors remain useful beyond their training scale, but editing additionally requires instruction fidelity and source preservation.

High-resolution editing follows two broad approaches. Cascades apply generative super-resolution after a 1K edit \citep{chen2025faithdiff,duan2025dit4sr,sun2025pixel,dong2025tsd}, reducing editor cost but moving detail reconstruction outside the editing trajectory. VINS-120K instead adapts the editor directly at 4K, retaining native denoising over long target--reference sequences \citep{chen2026vins}. Complementary methods address computational redundancy: RegionE uses adaptive region-wise denoising for editing, and Sparse-vDiT exploits structured attention sparsity in video diffusion Transformers \citep{chen2025regione,chen2026sparse}. Our focus is the cross-resolution coupling itself: a frozen 1K trajectory guides sparse, full-canvas target-resolution denoising within the same Transformer.

\section{Method}

We begin with flow-matching editing and its high-resolution attention bottleneck (\S\ref{sec:hr_edit}). We then adapt diffusion states and spatial positions to larger grids (\S\ref{sec:adapt}), introduce the sparse cross-resolution topology of \method{} (\S\ref{sec:flash}), and extend it with training-free progressive inference (\S\ref{sec:progressive}).

\begin{figure}[t]
\centering
\includegraphics[width=\columnwidth]{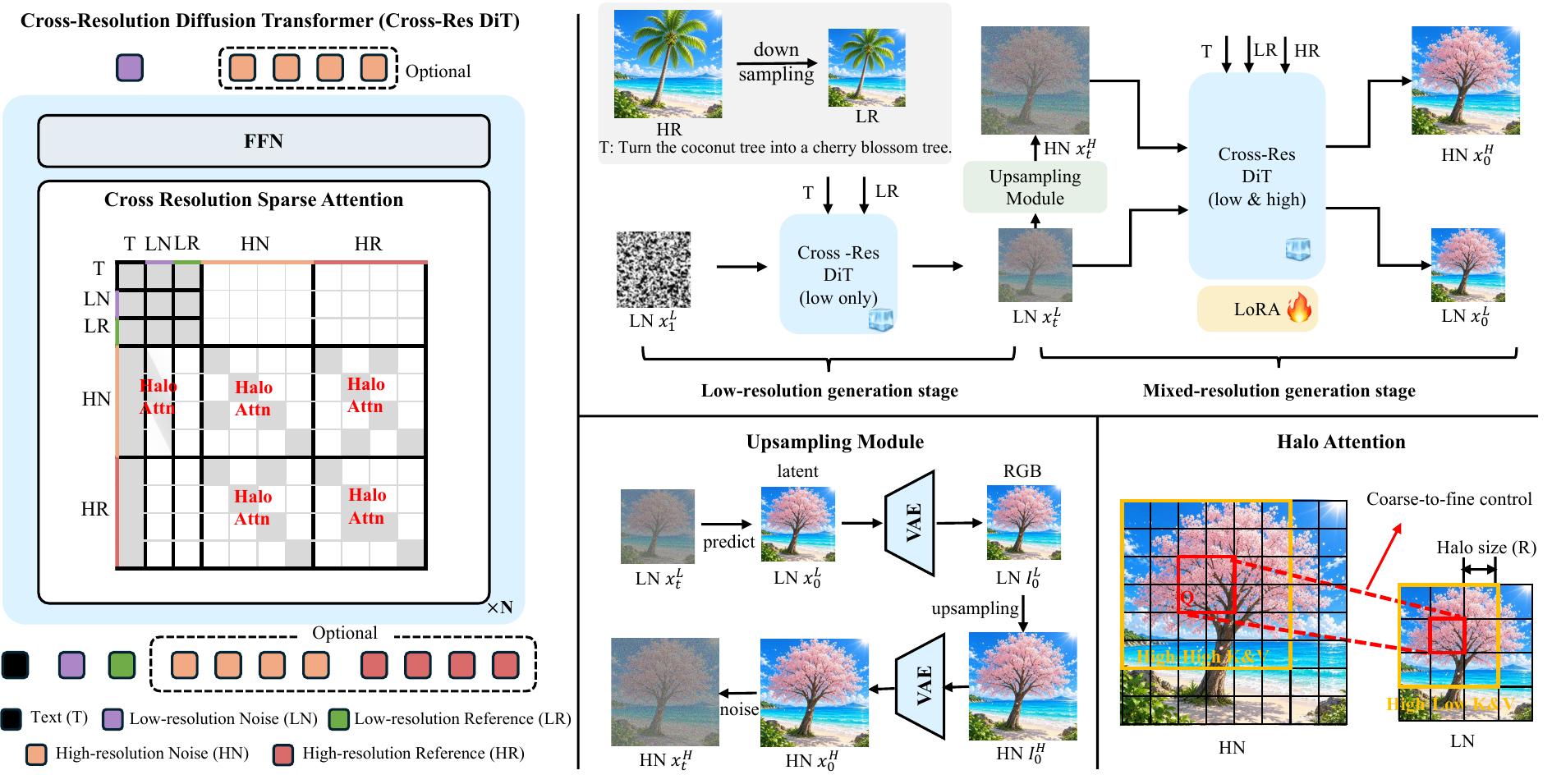}
\caption{Overview of \method{}. A frozen 1K branch anchors global editing,
local high-resolution attention reconstructs detail, and progressive inference
replaces early high-resolution steps with 1K denoising.}
\label{fig:main_method}
\end{figure}

\subsection{High-Resolution Image Editing}\label{sec:hr_edit}

\subsubsection{Flow-Matching Image Editing}

Given a source image \(I_R\) and instruction \(c\), the editor generates a
target \(I_0\) while preserving unrelated source content. The pretrained
variational autoencoder (VAE) encodes
\(I_R\) as clean conditioning latent \(z_R\), while the target latent starts
from Gaussian noise \(x_1\) \citep{rombach2022high}. The Transformer denoiser
predicts \(v_\theta(x_t,t;c,z_R)\) from noisy target, source, and text tokens;
integrating this field from \(t=1\) to \(0\) and decoding the final latent
yields the edit.

For training, encode \(I_0\) as clean latent \(z_0\), sample
\(t\in[0,1]\) and \(\epsilon\sim\mathcal N(0,I)\), and form the linear path
\[
x_t=(1-t)z_0+t\epsilon,
\qquad
v^\star=\frac{\mathrm d x_t}{\mathrm d t}=\epsilon-z_0.
\]

Here, \(t=0\) is the clean target and \(t=1\) is pure noise. We learn the conditional
velocity from \(c\) and \(z_R\) using the flow-matching objective \citep{lipman2022flow}
\[
\mathcal L_{\mathrm{FM}}
=
\mathbb E_{z_0,z_R,c,\epsilon,t}
\left[
\left\|
v_\theta(x_t,t;c,z_R)-(\epsilon-z_0)
\right\|_2^2
\right].
\]

\subsubsection{Computational Bottleneck at High Resolution}

With fixed latent downsampling and tokenization, scaling both image dimensions
by \(k\) increases spatial tokens by approximately \(k^2\). VAE, projections,
and FFNs grow with token count, whereas dense attention computes all
query--key pairs \citep{vaswani2017attention,crowson2024scalable}.
Memory-efficient exact attention reduces score materialization but retains
dense pairwise arithmetic \citep{dao2022flashattention}.
The independent \(R=1\) benchmark in Figure~\ref{fig:dense_scaling} illustrates
this bottleneck. Dense attention accounts for about 15.2 s of a
\(17.33\)-s 4K forward, compared with \(0.60\) s of a \(4.13\)-s forward
for the sparse diagnostic. The corresponding speedups are \(25.3\times\)
for attention and \(4.2\times\) overall. This diagnostic is separate from
the \(R=3\) quality comparison; Appendix~\ref{app:kernel} gives its settings.

\subsection{High-Resolution Capability Adaptation}\label{sec:adapt}

Applying a pretrained 1K editor to a larger grid introduces two mismatches. A nominal timestep can correspond to a different effective noise state, and spatial coordinates extend beyond the range seen during pretraining. We address the former with resolution-dependent timestep mapping and the latter with spatial YaRN.

\paragraph{Resolution-Dependent Timestep Mapping.}
Greater spatial redundancy at higher resolution can preserve more recoverable signal at the same nominal noise level. Reusing a lower-resolution schedule therefore changes the effective denoising difficulty. Resolution-dependent timestep shifts address this mismatch in high-resolution diffusion and rectified-flow models \citep{chen2023noise,hoogeboom2023simple,esser2024sd3}. Given a raw timestep \(u\), we map it to a resolution-specific state:
\[
s=f(u;\mu)
=
\frac{e^\mu u}
{1+(e^\mu-1)u},
\]
where \(\mu\) controls the shift. The raw timestep \(u\) corresponds to the nominal flow time \(t\) in Section~\ref{sec:hr_edit}; the shifted state \(s=f(u;\mu)\) is used both in \(x_s=(1-s)z_0+s\epsilon\) and as the timestep input to \(v_\theta\). At step \(k\), the branches share \(u_k\) but apply different shifts, \(s_{L,k}=f(u_k;\mu_L)\) and \(s_{H,k}=f(u_k;\mu_H)\). The paired states \((s_{L,k},s_{H,k})\) thus advance the two trajectories together while retaining their resolution-specific noise scales. Appendix~\ref{app:settings} gives the mapping parameters.

\paragraph{RoPE Position Adaptation.}
Applying a pretrained editor to a larger spatial grid requires rotary position
embedding (RoPE) to represent positions beyond the range observed during
pretraining
\citep{su2024roformer}. Insufficient adaptation can cause semantic drift or
local repetition in high-resolution editing \citep{chen2026vins}. Uniform
interpolation compresses phase differences across frequencies, so we use
YaRN's frequency-aware transition between interpolation and extrapolation
\citep{peng2024yarn}.

For a target-grid coordinate \(i\) relative to the shared spatial origin, the
pretrained-frame coordinate is \(p_i=i/n\), where \(n\) is the
target-to-pretrained scale. On a spatial axis with rotary dimension \(d_a\),
index \(j\) has native inverse frequency
\(\omega_j=\theta_{\mathrm{rope}}^{-2j/d_a}\), with rotary base
\(\theta_{\mathrm{rope}}\). YaRN adjusts it as
\[
\widetilde{\omega}_j
=
\omega_j\!\left[n-(n-1)\gamma_j\right],
\qquad
\phi_j(i)=p_i\widetilde{\omega}_j,
\]
where \(\gamma_j\in[0,1]\) is a linear ramp over rotary indices. Its endpoints
use rotation-count thresholds \((\alpha,\beta)=(1,8)\), computed from
\(r_j=L_{H,a}\omega_j/(2\pi)\), where \(L_{H,a}\) is the target-axis grid
length. High frequencies are extrapolated (\(\gamma_j\to0\)), while low
frequencies are fully interpolated (\(\gamma_j\to1\)). Non-spatial axes
retain native RoPE.

\subsection{Sparse Cross-Resolution Denoising}\label{sec:flash}

The adaptations above address diffusion states and spatial positions, but leave the dense-attention bottleneck intact. \method{} reduces this cost by coupling a global 1K editing anchor with local target-resolution detail generation. Here, \emph{cross-resolution} means jointly denoising both states within one Transformer, with explicitly constrained information flow rather than a post-hoc super-resolution bridge.

At each denoising step, our cross-resolution Transformer block jointly processes five token groups from a mixed-resolution token sequence,
\[
\mathcal{X}
=
[
\mathbf{T}_{\mathrm{text}},
\mathbf{T}_{L_N},
\mathbf{T}_{L_R},
\mathbf{T}_{H_N},
\mathbf{T}_{H_R}
],
\]
where \(\mathbf{T}_{\mathrm{text}}\) denotes text tokens; \(\mathbf{T}_{L_N}\) and \(\mathbf{T}_{L_R}\) denote the low-resolution noisy-target and clean-source tokens; and \(\mathbf{T}_{H_N}\) and \(\mathbf{T}_{H_R}\) denote their target-resolution counterparts. We use \emph{source} and \emph{reference} interchangeably for the conditioning image \(I_R\). This joint sequence is the basis of the attention topology below.

\vspace{-10pt}
\paragraph{Full-Attention 1K Anchor.}
The 1K branch retains the pretrained editor's native attention pattern, preserving instruction understanding, global layout, and the principal edit structure. Its image queries do not read high-resolution image keys, so attention from the high-resolution branch cannot change the 1K trajectory. This trajectory anchors target-resolution generation.

\vspace{-10pt}

\paragraph{Local High-Resolution Halo Attention.}
The high-resolution branch uses local target--source attention to recover native-grid structure and texture. Each low-resolution parent cell maps to an \(n\times n\) high-resolution region. Expanding this region by \(h\) latent-token units on each side gives its Halo support. We synchronize \(h\) with the low-resolution support radius \(R\) through \(h=Rn\), so both supports cover the same parent-cell neighborhood (\(h=6\) at 2K and \(h=12\) at 4K for \(R=3\)). Before boundary clipping, the support size per high-resolution segment is
\[
K_H=(n+2h)^2.
\]
The Halo contains both high-resolution noisy-target and reference tokens. It provides spatially aligned source details and exchanges context across neighboring parent cells, without dense attention over the complete target--source sequence. The support is fixed by local geometry and requires no additional user-provided edit mask.

\vspace{-10pt}

\paragraph{Spatially Aligned Low-to-High Target Guidance.}
Local Halo attention supplies target-scale context, but not the global edit established at 1K. The aligned pathway \(L_N\!\rightarrow H_N\) therefore carries the evolving low-resolution noisy-target state to high-resolution queries. With scale \(n\) along each spatial axis, a high-resolution location \((y,x)\) maps to the low-resolution parent cell
\[
\pi(y,x)
=
\left(
\left\lfloor\frac{y}{n}\right\rfloor,
\left\lfloor\frac{x}{n}\right\rfloor
\right).
\]
Each high-resolution noisy-target query reads the low-resolution noisy-target neighborhood of Chebyshev radius \(R\) around \(\pi(y,x)\), containing at most \(K_L=(2R+1)^2\) keys. Thus \(L_N\!\rightarrow H_N\) denotes information flowing from the low-resolution edit to the high-resolution target; computationally, high-resolution queries read low-resolution keys. This bounded pathway ties local refinement to the global editing trajectory.

For efficient execution, we group high-resolution tokens by their low-resolution parent cell. Contiguous parent-cell groups expose regular Halo and cross-resolution neighborhoods that the sparse-attention kernel evaluates without a dense mask or score matrix. The ordering preserves the attention topology while improving memory locality. Appendix~\ref{app:complexity} gives the support-count and cost derivations.

\paragraph{Training.}
The merged 1K weights are frozen and retain their native attention
connectivity. The branch remains in the forward pass to supply its evolving
editing state. Only the high-resolution branch is adapted, via rank-32
low-rank adaptation (LoRA) residuals on attention and feed-forward network
(FFN) projections. Training applies flow-matching regression to the
high-resolution velocity alone, without an auxiliary cross-resolution
consistency loss.

\subsection{Temporally Progressive Low-to-High-Resolution Editing}\label{sec:progressive}

Direct inference jointly denoises both branches over the full paired schedule. The training-free progressive path instead uses a 1K prefix for early, high-noise edit formation and reserves high-resolution computation for later refinement. For \(N\) denoising steps and transition index \(1\le K<N\), the model first runs the 1K segment, then uses the same cross-resolution model for the remaining trajectory.

At the transition, the low-resolution state \(x_L^K\) and its velocity prediction provide a clean estimate
\[
\hat{x}_{0,L}^{K}
=
x_L^K-s_{L,K}\,v_{\theta,L}^K .
\]
RGB-domain \emph{lifting} transfers this clean estimate to the target resolution:
\[
\hat I_L^K
=
D_{\mathrm{VAE}}\!\left(\hat{x}_{0,L}^{K}\right),
\qquad
\widetilde I_H^K
=
\mathcal R_{\mathrm{RGB}}\!\left(\hat I_L^K\right),
\hat{x}_{0,H}^{K}
=
E_{\mathrm{VAE}}\!\left(\widetilde I_H^K\right).
\]
Here \(D_{\mathrm{VAE}}\) and \(E_{\mathrm{VAE}}\) denote VAE decoding and
encoding, and \(\mathcal R_{\mathrm{RGB}}\) denotes RGB resizing to the
target resolution. The re-encoded state is then re-noised with independent
Gaussian noise at the high-resolution state paired with step \(K\):
\[
x_H^K
=
(1-s_{H,K})\hat{x}_{0,H}^{K}
+
s_{H,K}\epsilon_H,
\qquad
\epsilon_H\sim\mathcal N(0,I).
\]

The high-resolution branch then joins the low-resolution branch for the remaining paired steps. The RGB-resized image serves only as an initialization, not as the final output of an external super-resolution stage. The trained editor continues denoising under the instruction, source condition, and evolving low-resolution edit.

Increasing \(K\) reduces high-resolution computation but leaves fewer steps for target-scale refinement, directly controlling the quality--efficiency trade-off. We evaluate latent interpolation separately as a lifting ablation.

\section{Experiments}

\begin{table*}[t!]
\centering
\small
\setlength{\tabcolsep}{4.5pt}
\caption{Main 2K/4K comparison. Bold and underline mark the best and
second-best values per resolution; time is not ranked. HaarPSI uses the full
image. mPSNR/mSSIM/mMSE use the keep region; mLPIPS uses the edit region,
or the full image for global edits. Shaded rows do not use external
super-resolution. Time is end-to-end latency in seconds, excluding model
loading.}
\label{tab:main_2k4k}
\begin{tabular}{l|cccccc}
\toprule
\textbf{Method}
& \textbf{HaarPSI}\(\uparrow\)
& \textbf{mPSNR}\(\uparrow\)
& \textbf{mSSIM}\(\uparrow\)
& \textbf{mMSE}\(\downarrow\)
& \textbf{mLPIPS}\(\downarrow\)
& \textbf{Time}\(\downarrow\) \\
\midrule
\multicolumn{7}{c}{\textbf{2K Resolution}} \\
\midrule
FaithDiff~\citep{chen2025faithdiff} & 0.3840 & 21.7671 & 0.6900 & 0.01700 & 0.3101 & 29.43 \\
DiT4SR~\citep{duan2025dit4sr} & 0.3761 & 21.3047 & 0.6966 & 0.01757 & 0.3130 & 52.91 \\
PiSA-SR~\citep{sun2025pixel} & 0.3651 & 20.7503 & 0.6647 & 0.01843 & 0.3177 & 16.63 \\
TSD-SR~\citep{dong2025tsd} & 0.3600 & 20.1663 & 0.6468 & 0.02035 & 0.3338 & 16.89 \\
\rowcolor{gray!12}VINS~\citep{chen2026vins} & \underline{0.4164} & 22.2907 & 0.7065 & 0.01693 & \textbf{0.2486} & 100.82 \\
\rowcolor{gray!12}\method{} (Direct, \(R{=}3\)) & \textbf{0.4235} & \textbf{22.5668} & \textbf{0.7780} & \underline{0.01616} & \underline{0.2976} & 108.95 \\
\rowcolor{gray!12}\method{} (Progressive, \(R{=}3,K{=}12\)) & 0.4137 & \underline{22.4258} & \underline{0.7733} & \textbf{0.01554} & 0.3165 & 88.19 \\
\midrule
\multicolumn{7}{c}{\textbf{4K Resolution}} \\
\midrule
FaithDiff~\citep{chen2025faithdiff} & 0.3630 & 21.5728 & 0.6578 & 0.01741 & \underline{0.3079} & 87.04 \\
DiT4SR~\citep{duan2025dit4sr} & 0.3512 & 21.3398 & 0.6806 & 0.01786 & 0.3137 & 317.73 \\
PiSA-SR~\citep{sun2025pixel} & 0.3421 & 21.0066 & 0.6610 & 0.01817 & 0.3149 & 146.18 \\
TSD-SR~\citep{dong2025tsd} & 0.3352 & 20.5414 & 0.6329 & 0.01962 & 0.3161 & 62.46 \\
\rowcolor{gray!12}VINS~\citep{chen2026vins} & 0.2426 & 12.6890 & 0.4598 & 0.07081 & 0.4185 & 997.21 \\
\rowcolor{gray!12}\method{} (Direct, \(R{=}3\)) & \textbf{0.4067} & \textbf{22.2891} & \textbf{0.7621} & \textbf{0.01648} & \textbf{0.2934} & 309.69 \\
\rowcolor{gray!12}\method{} (Progressive, \(R{=}3,K{=}12\)) & \underline{0.3775} & \underline{21.6970} & \underline{0.7369} & \underline{0.01664} & 0.3190 & 215.31 \\
\bottomrule
\end{tabular}
\vspace{-10pt}
\end{table*}

We evaluate source-faithful editing and inference efficiency at 2K and 4K,
compare \method{} with edit--super-resolution cascades and native
high-resolution adaptation, and examine its design choices through ablations.
The supplementary material details mask construction and implementation.

\begin{figure}[t]
\centering
\includegraphics[width=\columnwidth]{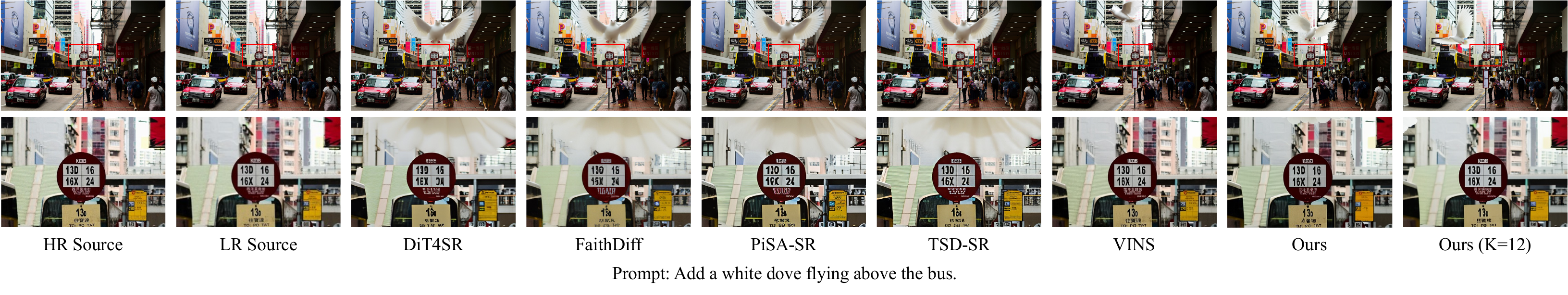}
\caption{2K outputs for ``Add a white dove flying above the bus.'' Direct and
progressive predictions are compared with edit--super-resolution cascades and
VINS; crops show the edit and nearby text detail.}
\label{fig:qualitative_main}
\end{figure}

\subsection{Experimental Settings and Evaluation Protocol}

We use FLUX.2 Klein Base 4B as the backbone, train on the full VINS-120K
split (131{,}010 examples), and evaluate on VINS-4KEval (509 instructions in
12 edit categories). All methods share source images, instructions, and
target resolutions. The 1K, 2K, and 4K settings use square
\(1024\times1024\), \(2048\times2048\), and \(4096\times4096\) canvases,
respectively. Local edits use a shared source-only mask; global edits are
excluded from keep-region scores and use the full image for mLPIPS. We
compare with FaithDiff, DiT4SR, PiSA-SR, TSD-SR, and training-free VINS.

The editing routes share a supervised fine-tuned 1K base. \method{} freezes
its low-resolution \(L=1\mathrm{K}\) branch and adapts
\(H\in\{2\mathrm{K},4\mathrm{K}\}\) with low-rank residuals \citep{hu2021lora}.
The main comparison uses \(R=3\), with \(K=12\) for progressive inference;
the ablations vary these design choices. Appendix~\ref{app:settings} gives
the adapter, topology, schedule, and optimizer settings.

We measure source fidelity with HaarPSI, a full-image Haar wavelet-based
perceptual similarity index, keep-region mPSNR/mSSIM/mMSE, and edit-region
mLPIPS. The \(m\) prefix denotes masking for local edits; global edits use
full-image LPIPS. These source--output metrics do not directly measure
instruction completion.
Quality values are macro averages over edit types. Timings exclude model
loading and include external super-resolution where applicable.

\subsection{Main Results}

The direct model leads the comparison in structural fidelity (Table~\ref{tab:main_2k4k}).
It has the highest HaarPSI, keep-region mPSNR, and mSSIM at both resolutions,
and the lowest 4K mMSE and source-relative mLPIPS. Progressive inference exchanges some
quality for lower latency: at 2K it stays within \(0.010\) HaarPSI and
\(0.005\) mSSIM of the direct model while reducing latency by \(19.1\%\).
At 4K it gives up \(0.029\) HaarPSI for a \(30.5\%\) latency reduction.

The separate forward diagnostic in Figure~\ref{fig:dense_scaling} evaluates
an \(R=1\) sparse configuration, not the \(R=3\) adapters in the main table.
It illustrates attention-cost reduction independently of the quality comparison.

\subsection{Ablation Studies}

\begin{table}[t]
\centering
\scriptsize
\setlength{\tabcolsep}{3.2pt}
\caption{2K synchronized Halo ablation. Increasing \(R\) expands low-
resolution support and the matched high-resolution Halo of width \(h=Rn\);
time is not ranked.}
\label{tab:ablation_halo}
\begin{tabular}{l|c|ccccc|c}
\toprule
\textbf{Support} & \(\boldsymbol{h}\)
& \textbf{HaarPSI}\(\uparrow\)
& \textbf{mPSNR}\(\uparrow\)
& \textbf{mSSIM}\(\uparrow\)
& \textbf{mMSE}\(\downarrow\)
& \textbf{mLPIPS}\(\downarrow\)
& \textbf{Time}\(\downarrow\) \\
\midrule
\(R=1\) & 2 & 0.4234 & 22.4356 & 0.7774 & 0.01591 & 0.2966 & 94.42 \\
\(R=2\) & 4 & 0.4213 & 22.4387 & 0.7732 & 0.01650 & 0.2977 & 99.64 \\
\(R=3\) & 6 & 0.4235 & 22.5668 & 0.7780 & 0.01616 & 0.2976 & 108.95 \\
\bottomrule
\end{tabular}
\end{table}

\begin{table*}[t]
\centering
\scriptsize
\setlength{\tabcolsep}{2.0pt}
\renewcommand{\arraystretch}{1.02}
\caption{Progressive transition ablation at \(R=3\). \(K\) is the initial
1K-prefix length; Direct denotes joint denoising. Time is end-to-end latency
in seconds, excluding model loading.}
\label{tab:ablation_transition}
\begin{minipage}[t]{0.49\textwidth}
\centering
\textbf{2K Resolution}\par\vspace{3pt}
\resizebox{\linewidth}{!}{%
\begin{tabular}{l|ccccc|c}
\toprule
\textbf{Transition}
& \textbf{HaarPSI}\(\uparrow\)
& \textbf{mPSNR}\(\uparrow\)
& \textbf{mSSIM}\(\uparrow\)
& \textbf{mMSE}\(\downarrow\)
& \textbf{mLPIPS}\(\downarrow\)
& \textbf{Time}\(\downarrow\) \\
\midrule
Direct & 0.4235 & 22.5668 & 0.7780 & 0.01616 & 0.2976 & 108.95 \\
\(K=8\) & 0.4173 & 22.4866 & 0.7768 & 0.01551 & 0.3132 & 102.32 \\
\(K=12\) & 0.4137 & 22.4258 & 0.7733 & 0.01554 & 0.3165 & 88.19 \\
\(K=16\) & 0.4097 & 22.4041 & 0.7693 & 0.01563 & 0.3157 & 74.84 \\
\bottomrule
\end{tabular}}
\end{minipage}
\hfill
\begin{minipage}[t]{0.49\textwidth}
\centering
\textbf{4K Resolution}\par\vspace{3pt}
\resizebox{\linewidth}{!}{%
\begin{tabular}{l|ccccc|c}
\toprule
\textbf{Transition}
& \textbf{HaarPSI}\(\uparrow\)
& \textbf{mPSNR}\(\uparrow\)
& \textbf{mSSIM}\(\uparrow\)
& \textbf{mMSE}\(\downarrow\)
& \textbf{mLPIPS}\(\downarrow\)
& \textbf{Time}\(\downarrow\) \\
\midrule
Direct & 0.4067 & 22.2891 & 0.7621 & 0.01648 & 0.2934 & 309.69 \\
\(K=8\) & 0.3888 & 21.8677 & 0.7470 & 0.01651 & 0.3118 & 257.22 \\
\(K=12\) & 0.3775 & 21.6970 & 0.7369 & 0.01664 & 0.3190 & 215.31 \\
\(K=16\) & 0.3682 & 21.5672 & 0.7299 & 0.01678 & 0.3205 & 174.12 \\
\bottomrule
\end{tabular}}
\end{minipage}
\end{table*}

\paragraph{Halo-Supported Local Attention.}

\begin{figure}[htbp]
\centering
\includegraphics[width=\columnwidth]{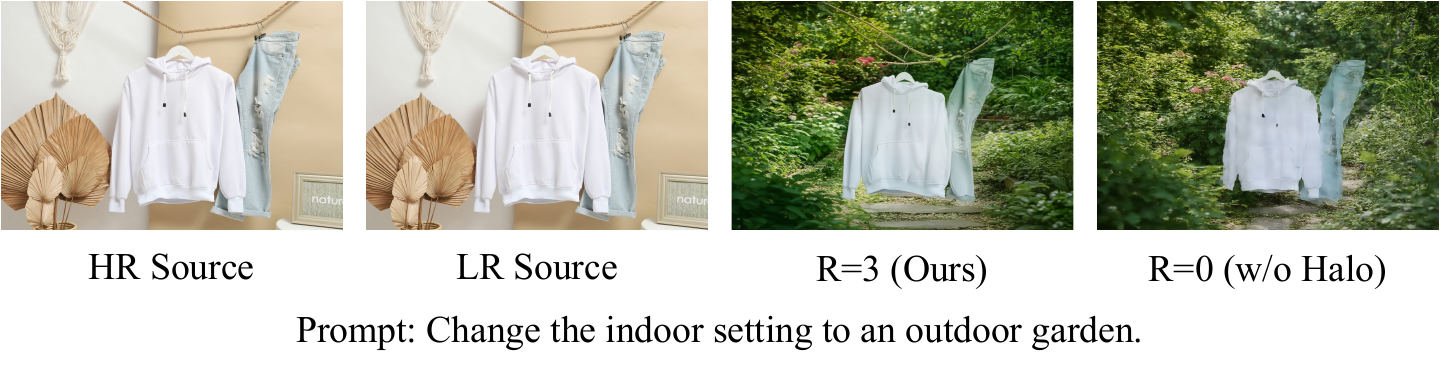}
\caption{2K Halo ablation under a shared instruction. \(R=3\) is compared
with no Halo \((R=0,h=0)\), highlighting cross-window continuity and texture.}
\label{fig:halo_ablation}
\end{figure}

We compare synchronized supports using a shared merged 1K base and
high-resolution adapters trained for the corresponding \(R\) values
(Table~\ref{tab:ablation_halo}).
At 2K, \(R=3\) gives the highest HaarPSI, mPSNR, and mSSIM, but also the
largest latency. The trend is not monotonic: \(R=2\) underperforms \(R=1\)
on most metrics, and \(R=1\) has the lowest keep-region mMSE and edit-region
mLPIPS. Thus a larger support does not improve every metric.
Figure~\ref{fig:halo_ablation} complements this comparison with a qualitative
\(R=0\) diagnostic: removing Halo support weakens cross-window continuity
and texture near the edit boundary.

\begin{figure}[htbp]
\centering
\includegraphics[width=\columnwidth]{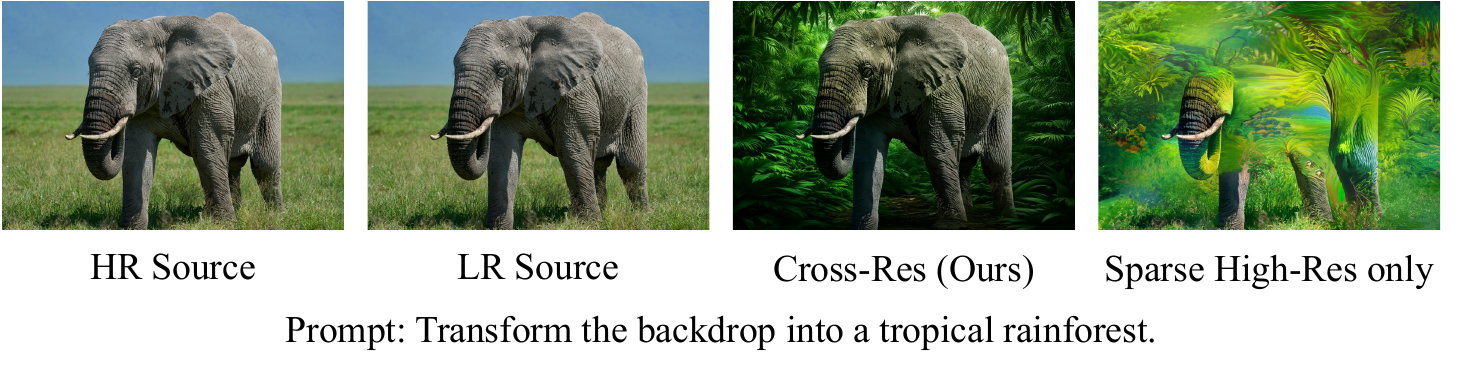}
\caption{2K anchor ablation. The anchor-free configuration removes
the 1K branch; the full model retains instruction-driven scene layout.}
\label{fig:hnln_ablation}
\end{figure}

\paragraph{Progressive Inference.}

Table~\ref{tab:ablation_transition} varies the transition index \(K\) at both
resolutions. A larger \(K\) replaces more high-resolution steps with the 1K
prefix, leaving fewer steps for target-scale refinement. At 2K, \(K=8\)
gives the strongest progressive quality among the tested transitions. At
4K, latency reductions increase with \(K\); we use \(K=12\) in the main
comparison. The table reports end-to-end timings for \(K=8,12,16\).

\paragraph{Low-Resolution Anchor and Low-to-High Guidance.}

We remove the complete 1K anchor while retaining the high-resolution sparse
branch. This also removes the evolving low-resolution target state and its
aligned guidance, leaving bounded high-resolution interactions to infer
instruction-conditioned layout and object placement. Figure~\ref{fig:hnln_ablation}
shows the resulting structural failure mode. Appendix~\ref{app:topology}
documents the full-model topology, including the removed
\(L_N\!\rightarrow H_N\) pathway.

\begin{figure}[htbp]
\centering
\includegraphics[width=\columnwidth]{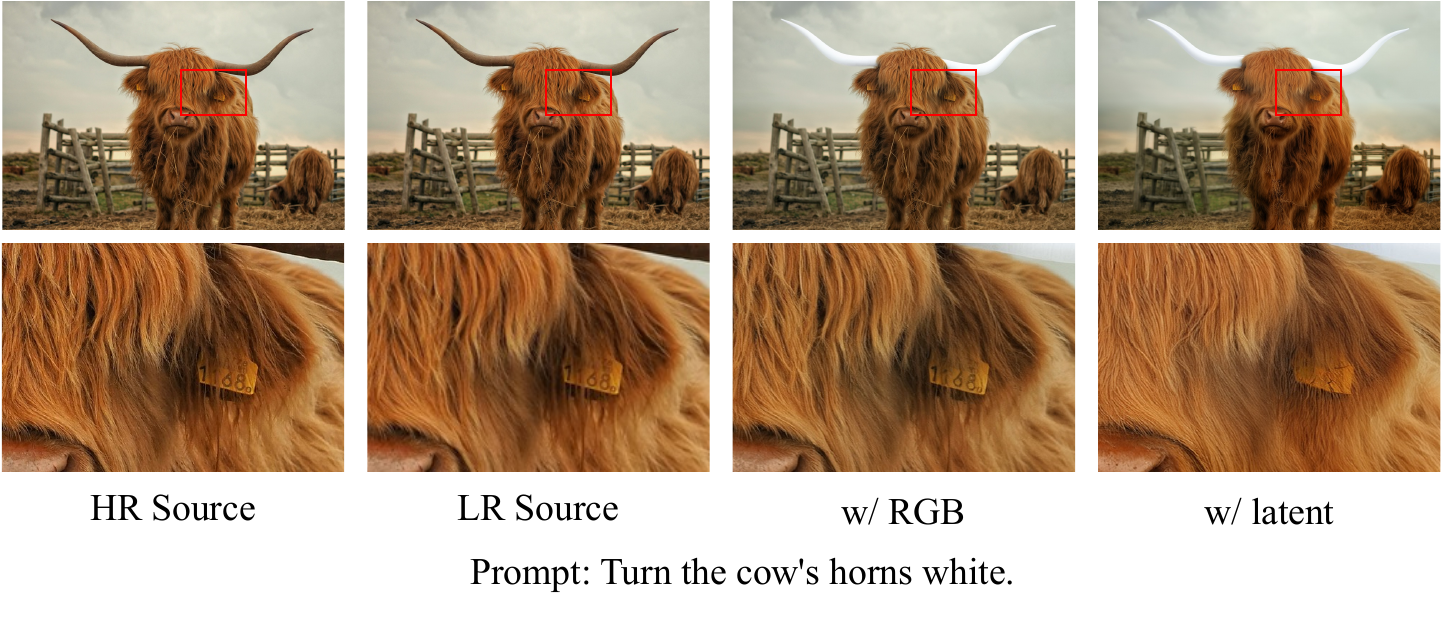}
\caption{2K lifting ablation at \(R=3\). RGB/VAE and latent configurations share
the same progressive denoising and differ only in the lifting operator.}
\label{fig:lifting_ablation}
\end{figure}

\paragraph{RGB/VAE versus Latent Lifting.}
We compare RGB resizing followed by target-resolution VAE encoding with
direct latent interpolation, keeping subsequent denoising identical.
Figure~\ref{fig:lifting_ablation} isolates the lifting operator and shows
better detail preservation with RGB-space upsampling.

\section{Conclusion}

We presented \method{}, a native high-resolution editor that couples a frozen 1K editing anchor with sparse target-resolution refinement. Aligned low-to-high guidance connects the global editing trajectory to local Halo attention, while resolution-specific diffusion states and spatial position adaptation extend the pretrained editor to larger grids. Optional progressive inference replaces early high-resolution steps with a 1K prefix, controlling the quality--latency trade-off through the transition index. Experiments at 2K and 4K demonstrate this framework as a practical alternative to dense native editing and edit--then--refine cascades.

\subsection*{AI Use Statement}

AI tools assisted with manuscript editing and preparation of supporting
release code. The authors are responsible for reviewing these contributions
and for the accuracy of the manuscript, experimental results, figures, and
released artifacts.

\subsection*{Ethics Statement}

This work does not involve human subjects, new personal-data collection, or
human-participant experiments. We evaluate image editing models on an existing
benchmark and report model behavior, source-image preservation, and
computational cost. The released evaluation protocol should be used in
accordance with the licenses and terms of the underlying benchmark and model
checkpoints.

\subsection*{Reproducibility Statement}

The main paper describes the model, token topology, resolution mapping,
training objective, and evaluation metrics. The supplementary material gives
training and inference settings (Appendix~\ref{app:settings}), algorithms
(Appendix~\ref{app:pseudocode}), the evaluation-mask protocol
(Appendix~\ref{app:metrics}), topology and kernel details
(Appendices~\ref{app:topology} and~\ref{app:kernel}), and additional
visualizations (Appendix~\ref{app:visuals}). Reproduction requires the merged
1K checkpoint, its matching high-resolution adapter, the settings in
Appendix~\ref{app:settings}, and the evaluated connectivity in
Appendix~\ref{app:topology}.

\bibliographystyle{iclr2027_conference}
\bibliography{references}

\newpage



\clearpage
\appendix
\centering \textbf{\large Efficient and Faithful High-Resolution Image Editing with Sparse Cross-Resolution Denoising}

\vspace{5pt}
\centering {\large Supplementary Material}

\justify

\vspace{10pt}

\noindent
This supplement details training and inference settings, algorithms,
evaluation masks and metrics, attention topology, kernel implementation,
complexity, limitations, and additional visual comparisons.

\section{Training and Inference Settings}
\label{app:settings}

\subsection{Model initialization and data}

We instantiate \method{} on FLUX.2 Klein Base 4B. We first fine-tune the 1K
editor on the VINS-120K split (131{,}010 examples), merge it into the base
Transformer, and freeze it. High-resolution adaptation then trains only a
high-resolution LoRA branch; it neither re-trains the text encoder nor
replaces the timestep embedding network.

The targets \(H\in\{2\mathrm{K},4\mathrm{K}\}\) are square
\(2048\times2048\) and \(4096\times4096\) canvases. With \(L=1\mathrm{K}\)
and \(n=H/L\), each low-resolution parent cell corresponds to an
\(n\times n\) group of high-resolution latent tokens. Target and reference
images are encoded natively at each resolution; the high-resolution noisy
latent is not obtained by upsampling a low-resolution noisy latent.

\subsection{Trainable parameters and optimization}

High-resolution adaptation uses rank-32 LoRA on attention and FFN
projections. The 1K adapter has already been merged into the frozen base;
only the high-resolution residual path is updated.

All high-resolution variants use AdamW with the settings in
Table~\ref{tab:settings}. The initial 2K screening stage has a matched
2{,}000-step budget, with the same data order, seed policy, adapter rank,
optimizer, and inference protocol across variants. Each run archives its
checkpoint, sampler, data split, step count, and evaluation set.

\begin{table}[h]
\centering
\small
\caption{Settings used in this work.}
\label{tab:settings}
\begin{tabular}{@{}p{0.30\columnwidth}p{0.58\columnwidth}@{}}
\toprule
\textbf{Item} & \textbf{Setting} \\
\midrule
Base model & FLUX.2 Klein Base 4B \\
Low-resolution editor & 1K supervised fine-tuned checkpoint, merged and frozen \\
Training data & Full VINS-120K split; 131{,}010 examples \\
High-resolution branch & Rank-32 LoRA on attention and FFN projections \\
Optimizer & AdamW; learning rate \(2\times10^{-5}\); weight decay \(0.01\) \\
Warm-up & 100 optimizer steps \\
High-resolution targets & \(2\mathrm{K}\) and \(4\mathrm{K}\) \\
Low-to-high scale & \(n=2\) for 2K; \(n=4\) for 4K \\
RoPE & YaRN on the two spatial axes; image-stream and text axes keep native RoPE; rotation-count thresholds \((\alpha,\beta)=(1,8)\) \\
Timestep conditioning & Paired low/high branch conditioning with shared raw timesteps \\
Sparse support & \(R=3\) with synchronized Halo \(h=Rn\) \\
\bottomrule
\end{tabular}
\end{table}

\subsection{Noise schedule and conditioning}

The training sampler draws a shared raw \(u\in[0,1]\) and applies the
resolution-dependent flow-matching shift
\[
s(u;\mu)=\frac{\exp(\mu)u}{1+(\exp(\mu)-1)u}.
\]
The low-resolution branch follows the native 1K schedule. The high-resolution
branch uses its paired state from Section~\ref{sec:adapt} and receives its own
timestep in the branch-conditioned Transformer blocks; reference rows remain
clean. For FLUX.2, \(\mu_L=1.15\), with \(\mu_H=1.380517\) at 2K and
\(\mu_H=2.302585\) at 4K.

The training loss is the flow-matching velocity regression on native
resolution latents. For a clean latent \(z_0\), independent Gaussian noise
\(\epsilon\), and branch state \(s\),
\[
x_s=(1-s)z_0+s\epsilon,
\qquad
v^\star=\epsilon-z_0.
\]
Only the high-resolution branch is optimized against \(v^\star_H\); the frozen
1K branch supplies the editing anchor. No auxiliary cross-resolution
consistency loss is used.

\subsection{Direct and progressive inference}

Direct inference uses 28 denoising steps, classifier-free guidance 4.0, and
one image per batch along the full joint trajectory. It applies YaRN RoPE
and disables KV caching. Timing excludes one-time model loading; for the
progressive path it includes the complete 1K prefix and the resolution
transition.

Progressive inference uses \(K=12\) in the main comparison. After the first
\(K\) native 1K steps, it forms a clean estimate, decodes it to RGB, resizes
and re-encodes it, then re-noises it at the paired high-resolution state.
The remaining \(N-K\) steps use the direct cross-resolution model. Latent
interpolation is evaluated separately as a lifting ablation.

\section{Training and Inference Algorithms}
\label{app:pseudocode}

\begin{samepage}
\subsection{High-resolution training}

\begin{algorithm}[H]
\caption{High-resolution branch training}
\label{alg:training}
\begin{algorithmic}[1]
\Require source/target images $I_R$, $I_0$; instruction $c$; frozen 1K editor $L$ and trainable high-resolution LoRA parameters $\theta_H$
\For{each training example}
    \State $(\mathbf{z}_L^R,\ \mathbf{z}_L^0) \gets \Call{VAE\_encode}{I_R \text{ at } 1\mathrm{K}},\ \Call{VAE\_encode}{I_0 \text{ at } 1\mathrm{K}}$
    \State $(\mathbf{z}_H^R,\ \mathbf{z}_H^0) \gets \Call{VAE\_encode}{I_R \text{ at } H},\ \Call{VAE\_encode}{I_0 \text{ at } H}$
    \State $u \gets \operatorname{Uniform}(0,1)$;\quad $s_L \gets \operatorname{shift}(u;\mu_L)$;\quad $s_H \gets \operatorname{shift}(u;\mu_H)$
    \State $\boldsymbol{\epsilon}_L, \boldsymbol{\epsilon}_H \gets$ independent standard Gaussian noise
    \State $\mathbf{x}_L \gets (1-s_L)\,\mathbf{z}_L^0 + s_L\,\boldsymbol{\epsilon}_L$;\quad $\mathbf{x}_H \gets (1-s_H)\,\mathbf{z}_H^0 + s_H\,\boldsymbol{\epsilon}_H$
    \State $\mathbf{X} \gets \Call{pack}{\operatorname{text}(c),\ \mathbf{x}_L,\ \mathbf{z}_L^R,\ \mathbf{x}_H,\ \mathbf{z}_H^R}$
    \State $\mathbf{X}_H \gets$ high-resolution segment of $\mathbf{X}$
    \State $\mathbf{X}_H \gets \Call{parent\_cell\_permutation}{\mathbf{X}_H,\ n}$
    \State $\mathbf{X} \gets$ replace high-resolution segment in $\mathbf{X}$ with $\mathbf{X}_H$
    \State $(\mathbf{v}_L,\ \mathbf{v}_H) \gets \Call{sparse\_cross\_resolution\_transformer}{\mathbf{X},\ s_L,\ s_H;\ \theta_H}$
    \State $\mathcal{L} \gets \operatorname{mean\_square}(\mathbf{v}_H,\ \boldsymbol{\epsilon}_H - \mathbf{z}_H^0)$
    \State update $\theta_H$ using AdamW($\mathcal{L}$)
\EndFor
\State \Return frozen merged 1K editor and high-resolution LoRA checkpoint
\end{algorithmic}
\end{algorithm}
\end{samepage}

The low-resolution branch is present in the forward pass so that the
high-resolution branch can read the evolving editing state. Freezing the
merged 1K branch prevents high-resolution adaptation from changing the
native 1K editing prior.

\begin{samepage}
\subsection{Parent-cell permutation}

\begin{algorithm}[H]
\caption{Window-major parent-cell permutation}
\label{alg:permutation}
\begin{algorithmic}[1]
\Require low and high token-grid dimensions $(H_L, W_L)$ and $(H_H, W_H)$; scale $n = H_H/H_L$
\For{each high token coordinate $(y, x)$}
    \State $p_y \gets \lfloor y/n \rfloor$;\quad $p_x \gets \lfloor x/n \rfloor$;\quad $\ell_y \gets y \bmod n$;\quad $\ell_x \gets x \bmod n$
    \State $key \gets (p_y,\ p_x,\ \ell_y,\ \ell_x)$
    \State append original index to $bucket[key]$
\EndFor
\State $perm \gets$ concatenate buckets in parent-cell row-major and local row-major order
\State $\mathbf{X}_{perm} \gets \Call{gather}{\mathbf{X}_{high},\ perm}$
\State $meta_{perm} \gets \Call{gather}{token\_metadata,\ perm}$
\State $inverse\_perm \gets \Call{invert\_permutation}{perm}$
\State \Return $\mathbf{X}_{perm},\ meta_{perm},\ inverse\_perm$
\end{algorithmic}
\end{algorithm}
\end{samepage}

The permutation makes all \(n^2\) high-resolution tokens belonging to the
same low-resolution parent cell contiguous. It changes storage order only;
the inverse permutation restores the canonical sequence before decoding or
writing diagnostics.

\begin{samepage}
\subsection{Sparse attention support}

\begin{algorithm}[H]
\caption{Support construction for one Transformer block}
\label{alg:support}
\begin{algorithmic}[1]
\Require token metadata; low-resolution radius $R$; Halo width $h$; enabled cross-resolution paths
\For{each query $q$}
    \If{$q$ is text or a low-resolution image token}
        \State $keys(q) \gets$ text, \(\LN\), and \(\LR\) tokens (native global support)
    \ElsIf{$q$ is $H_N$}
        \State $p \gets \Call{parent\_cell}{q}$
        \State $K_{hi} \gets \HN$ and $\HR$ tokens in the local Halo around $p$
        \State $K_{lo} \gets \LN$ tokens in the parent neighborhood of Chebyshev radius $R$
        \State $K_{txt} \gets$ text tokens
        \State $keys(q) \gets K_{hi} \cup K_{lo} \cup K_{txt}$
    \ElsIf{$q$ is $H_R$}
        \State $keys(q) \gets$ permitted text tokens plus the local high-resolution support
    \EndIf
    \State emit row offsets and exact key indices for $q$
\EndFor
\State apply one normalized sparse softmax over each emitted row
\State \Return sparse attention output and row metadata
\end{algorithmic}
\end{algorithm}
\end{samepage}

The high-resolution query rows are formed from exact spatial indices rather
than a dense mask, and boundary cells are clipped to the valid image grid.
Training evaluates the same connectivity with a dense block-sparse mask, while
inference uses the compressed-sparse-row (CSR) kernel described in
Section~\ref{app:kernel}. In the
configuration used throughout, high-resolution queries do not read
low-resolution reference keys, while only \(\HN\) queries retain the local
low-resolution noisy-target support.

\begin{samepage}
\subsection{Progressive low-to-high inference}

\begin{algorithm}[H]
\caption{Training-free progressive inference}
\label{alg:progressive}
\begin{algorithmic}[1]
\Require source image $I_R$, instruction $c$, total steps $N$, transition index $K$
\State $\mathbf{z}_L^R \gets \Call{VAE\_encode}{I_R \text{ at } 1\mathrm{K}}$;\quad $\mathbf{z}_H^R \gets \Call{VAE\_encode}{I_R \text{ at target resolution } H}$
\State $\mathbf{x}_L \gets \Call{initialize\_noise}{1\mathrm{K}}$
\For{$i = 0, \dots, K-1$}
    \State $\mathbf{v}_L \gets \Call{frozen\_1K\_editor}{\mathbf{x}_L,\ \mathbf{z}_L^R,\ c,\ s_L[i]}$
    \State $\mathbf{x}_L \gets \Call{Euler\_step}{\mathbf{x}_L,\ \mathbf{v}_L,\ s_L[i] \to s_L[i{+}1]}$
\EndFor
\State $\mathbf{v}_L \gets \Call{frozen\_1K\_editor}{\mathbf{x}_L,\ \mathbf{z}_L^R,\ c,\ s_L[K]}$
\State $\mathbf{x}_{0,L} \gets \Call{clean\_estimate}{\mathbf{x}_L,\ \mathbf{v}_L,\ s_L[K]}$
\State $\hat{I}_L \gets \Call{VAE\_decode}{\mathbf{x}_{0,L}}$
\State $\tilde{I}_H \gets \Call{rgb\_resize}{\hat{I}_L,\ \text{target resolution } H}$
\State $\mathbf{x}_{0,H} \gets \Call{VAE\_encode}{\tilde{I}_H}$
\State $\mathbf{x}_H \gets (1-s_H[K])\,\mathbf{x}_{0,H} + s_H[K]\,\Call{Gaussian\_noise}{}$
\For{$i = K, \dots, N-1$}
    \State $(\mathbf{v}_L,\ \mathbf{v}_H) \gets \Call{cross\_resolution\_model}{\mathbf{x}_L,\ \mathbf{x}_H,\ \mathbf{z}_L^R,\ \mathbf{z}_H^R,\ c,\ s_L[i],\ s_H[i]}$
    \State $(\mathbf{x}_L,\ \mathbf{x}_H) \gets \Call{paired\_Euler\_step}{\mathbf{x}_L,\ \mathbf{x}_H,\ \mathbf{v}_L,\ \mathbf{v}_H}$
\EndFor
\State \Return $\Call{VAE\_decode}{\mathbf{x}_H}$
\end{algorithmic}
\end{algorithm}
\end{samepage}

\section{Evaluation Masks and Metrics}
\label{app:metrics}

\subsection{Evaluation protocol}

VINS-4KEval contains 509 instructions across 12 edit categories. Before
comparing methods, we generate each mask from the source image and the
instruction-derived edit specification only. Outputs are never used in mask
construction. Every method, resolution route, and transition uses the same
mask for a given case.

We use GroundingDINO-Tiny in FP32 and SAM2.1 Hiera Tiny in the model dtype.
Detection thresholds are \(0.25\) for boxes and \(0.20\) for text, with at
most four detections per query. An elliptical structuring element dilates
the mask with a radius of \(1.5\%\) of the shorter source-image dimension.

\subsection{Instruction parsing and query routing}

A deterministic parser extracts the mask scope, source and requested object
phrases, spatial anchor and relation, and parsing status. Where possible, part-level requests
such as ``eyes,'' ``beak,'' or ``horns'' are queried directly rather than
expanded to the whole object. Replacement records both object phrases;
addition uses a source-only anchor because the requested object is absent.

\subsection{Mask construction}

For local deletion, replacement, and attribute edits, GroundingDINO detects
the source phrase and passes its boxes to SAM2. We take the mask union as
\(M_e\) and apply the fixed dilation. If SAM2 is unavailable, detector boxes
provide a marked coarse fallback, kept separate from SAM2 masks.

For addition, relations such as ``above,'' ``near,'' ``left,'' and ``in the
background'' determine a region around the detected source anchor. If no
concrete anchor is detected, a deterministic foreground, background, left,
or right region is selected from the instruction. The added object is never
localized from the generated image.

For background edits, the evaluator detects a category-aware source
foreground query (for example, \texttt{animal.}, \texttt{person.},
\texttt{vehicle.}, or \texttt{object.}) and segments it with SAM2. Let
\(M_f\) be the dilated foreground mask. The editable region is
\[
M_e = 1-M_f,
\qquad
M_k=M_f,
\]
so the changed background is scored as editable while the source foreground
is evaluated for preservation. Global tone, style, camera, and personalized
generation edits use \(M_e=\mathbf{1}\) and are excluded from local
keep-region metrics.

\subsection{Resolution normalization and per-image metrics}

Every output is converted to the common target canvas before mask application.
The mask is resized with nearest-neighbor interpolation. Let \(I_s\) be the
source, \(I_o\) the output, \(M_e\) the editable region, and
\(M_k=1-M_e\) the keep region.

\paragraph{Non-edited-region preservation.}
For a valid keep region \(M_k\), let \(I_s(p),I_o(p)\in[0,255]^3\) be the
8-bit RGB vectors at pixel \(p\). We average over pixels and channels and
normalize the squared error by \(255^2\):
\[
\operatorname{mMSE}
=
\frac{1}{3\,255^2|M_k|}\sum_{p\in M_k}
\left\|I_s(p)-I_o(p)\right\|_2^2.
\]
The corresponding per-image PSNR is
\[
\operatorname{mPSNR}=-10\log_{10}\operatorname{mMSE}.
\]
For zero squared error, mPSNR is \(+\infty\). These metrics are computed per
image before macro-averaging; the mean PSNR is not inferred from the mean MSE.
SSIM is computed after replacing the masked edit region with the same neutral
value in both images. The keep-region mask is eroded by the SSIM window
radius before averaging the SSIM map, with identical handling for all
methods.

\paragraph{Edit-region perceptual distance.}
For local edits, mLPIPS is the source--output perceptual distance on the
editable region. Before compositing, images are resized to a maximum long
side of 1024 pixels with Lanczos interpolation; masks use nearest-neighbor
resizing. Both images are composited against the same neutral background
outside \(M_e\), and LPIPS is computed on the pair,
\[
\operatorname{mLPIPS}
=
\operatorname{LPIPS}\!\left(
\operatorname{comp}(I_s,M_e),
\operatorname{comp}(I_o,M_e)
\right).
\]
For global edits, \(M_e=\mathbf{1}\), so this score is computed on the full
source--output pair. It measures perceptual change from the source, not
instruction completion. Keep-region and edit-region scores are aggregated
separately.

\paragraph{Full-image structural score.}
\looseness=-1
HaarPSI is computed on the full normalized source/output pair, without
restriction to \(M_e\) or \(M_k\).

\subsection{Aggregation and validity}

Per-sample records include the case, method, edit type, mask scope and
coverage, normalization resolution, and metric values. We first average
within each edit type over applicable finite scores and then across types,
so larger categories do not dominate. Global edits contribute to full-image metrics but not local
masked-preservation averages. Detector-based edit-success flags remain
automatic proxies rather than human-verified semantic judgments.

We retain source queries, boxes, confidence scores, mask coverage, and visual
sheets with the source, shared mask, and outputs with mask overlays. Ambiguous part-level
masks are manually overridden or excluded, with the reason recorded. OCR
exact-match and CER are not reported.

\section{Low--High Attention Topology}
\label{app:topology}

\subsection{Information-flow statement}

Arrows \(A\rightarrow B\) denote information flow, not query--key matrix
direction. The evaluated topology retains one direct cross-resolution
guidance path:
\[
\LN\longrightarrow\HN.
\]
Its implementation is a high-resolution noisy-target query reading
low-resolution noisy-target keys and values:
\[
\operatorname{Attn}_{\LN\rightarrow\HN}
\left(
Q_{\HN},K_{\LN},V_{\LN}
\right).
\]
Thus high-resolution queries receive guidance from the evolving 1K edit.

\subsection{What is retained}

The low-resolution branch keeps its native global editing interaction over
text, \(\LN\), and \(\LR\). High-resolution noisy-target queries use a
spatially aligned neighborhood of \(\LN\) keys, where the parent cell of a
high-resolution query is
\[
\parent(y,x)
=
\left(
\left\lfloor y/n\right\rfloor,
\left\lfloor x/n\right\rfloor
\right).
\]
The support is bounded by a Chebyshev radius \(R\), giving at most
\[
K_L=(2R+1)^2
\]
low-resolution target keys before image-boundary clipping. High-resolution
target/reference tokens interact locally through the Halo support on the
high-resolution grid. These are local same-resolution source/detail
interactions, not additional cross-resolution reference bridges.

\subsection{What is not retained}

In the evaluated configuration, high-resolution queries exclude low-resolution
reference keys. \(H_N\) queries still read their local \(\LN\) neighborhood.
The low-resolution reference remains available to the 1K editor, and the
high-resolution reference enters through local Halo support.

\subsection{Query--key adjacency in the reported configuration}

Table~\ref{tab:adjacency} gives the full query--key connectivity. Text and
low-resolution rows attend globally to Text, \(\LN\), and \(\LR\).
\(\HN\) rows read text, their aligned \(\LN\) neighborhood, and local Halo
keys; \(\HR\) rows read text and local Halo keys. Each row uses one attention
distribution over the union of admitted keys. High-resolution rows do not
read low-resolution reference keys.

\begin{table}[h]
\centering
\small
\caption{Query--key adjacency of the evaluated configuration. Local denotes the
Chebyshev-\(R\) parent-cell neighborhood (\(K_L=(2R+1)^2\) keys) or the Halo
support (\(K_H=(n+2h)^2\) keys); \checkmark{} denotes the native global
pattern.}
\label{tab:adjacency}
\begin{tabular}{l|ccccc}
\toprule
Query $\backslash$ Key & Text & $\LN$ & $\LR$ & $\HN$ & $\HR$ \\
\midrule
Text & \checkmark & \checkmark & \checkmark & -- & -- \\
$\LN$ & \checkmark & \checkmark & \checkmark & -- & -- \\
$\LR$ & \checkmark & \checkmark & \checkmark & -- & -- \\
$\HN$ & \checkmark & local & -- & local & local \\
$\HR$ & \checkmark & -- & -- & local & local \\
\bottomrule
\end{tabular}
\end{table}

\subsection{Scope of the comparison}

The topology retains \(\LN\rightarrow\HN\) guidance and excludes
low-resolution reference keys from high-resolution queries. These choices
apply to the evaluated configuration, not to every reference-fusion design.

\section{Sparse Kernel Implementation}
\label{app:kernel}

\subsection{Sequence layout and metadata}

The canonical packed sequence is
\[
[\mathbf{T}_{\mathrm{text}},
\mathbf{T}_{\LN},
\mathbf{T}_{\LR},
\mathbf{T}_{\HN},
\mathbf{T}_{\HR}].
\]
Before attention, high-resolution tokens are reordered by parent cell.
Permuted metadata records segment identity, parent-cell and high latent-token
coordinates, and inverse-permutation indices. The kernel uses these indices
to construct exact supports without a dense \((S\times S)\) mask.

\subsection{CSR attention kernel}

For 2K inference, text and low-resolution rows use FlashAttention, while
high-resolution rows use a Triton CSR kernel over paired high-resolution
target/reference rows with
\[
R=3,\qquad h=6\;(2\mathrm{K}).
\]
Each descriptor stores valid key indices and offsets, avoiding scores for
masked keys. Boundary rows are clipped to their valid prefix. Training uses
a dense block-sparse mask with the same connectivity.

High-resolution support combines local target/reference keys, text, and the
aligned \(\LN\) neighborhood for \(\HN\) queries. A single attention
distribution normalizes this union.

\subsection{Correctness and runtime safeguards}

We validate each permutation and CSR builder against a dense reference: the
densified support must match the intended topology, and the inverse
permutation must restore the original token order. Boundary cells are checked
separately from interior parent cells because their valid key counts differ.
Each run records model-loading time, generation time, and peak memory.

\subsection{Independent forward diagnostic}

Figure~\ref{fig:dense_scaling} reports a separate single-forward diagnostic.
The dense comparator processes text and target/reference image tokens without
a low-resolution image branch. The sparse configuration uses \(R=1\),
\(h=2\) at 2K and \(h=4\) at 4K, with high-resolution shifts \(1.380517\)
and \(3.229986\), respectively. These are the benchmark settings, not the
\(R=3\), \(h=6/12\) quality configuration. Its \(4.2\times\) total-forward
and \(25.3\times\) attention ratios therefore characterize that diagnostic
only, not end-to-end generation or progressive inference.

\section{Complexity Derivation}
\label{app:complexity}

\subsection{Notation}

Let \(H_L,W_L\) and \(H_H,W_H\) be the low- and high-resolution
token-grid dimensions, and let \(P_L=H_LW_L\) and \(P_H=H_HW_H=n^2P_L\) be
the corresponding image-token counts per image stream. Let \(T\) be the text
token count, \(d\) the head dimension, \(R\) the low-resolution support
radius, and \(h\) the high-resolution Halo width. The high-resolution local
support before boundary clipping contains
\[
K_H=(n+2h)^2
\]
tokens per image stream. The aligned low-resolution target support contains
\(K_L=(2R+1)^2\) keys.

\subsection{Dense attention}

The native dense comparator in Figure~\ref{fig:dense_scaling} uses a
sequence of length \(T+2P_H\), without low-resolution image tokens. For
comparison with a fully dense mixed-resolution topology, all five token
groups would participate in global attention, giving
\[
S_{\mathrm{dense}}=T+2P_L+2P_H.
\]
The score and value-mixing work of one dense attention layer is
\[
\mathcal{C}_{\mathrm{dense}}
=
\Theta\!\left(S_{\mathrm{dense}}^2d\right),
\qquad
\mathcal{M}_{\mathrm{dense}}
=
\Theta\!\left(S_{\mathrm{dense}}^2\right)
\]
if the score matrix is materialized. Flash-style kernels reduce score
storage, but the dense pairwise arithmetic remains quadratic in
\(S_{\mathrm{dense}}\).

\subsection{Sparse cross-resolution attention}

The low-resolution/text part retains native global interaction:
\[
\mathcal{C}_{L}
=
\Theta\!\left((T+2P_L)^2d\right).
\]
For high-resolution query rows, each query sees a finite support and, for
\(\HN\) rows, an additional low-resolution target neighborhood. Writing
\(Q_H\) for the number of high-resolution query rows gives the conservative
upper bound
\[
\mathcal{C}_{H}
=
O\!\left(
Q_H
\left[
T+2(n+2h)^2+(2R+1)^2
\right]d
\right).
\]
For the common paired target/reference layout \(Q_H\) is proportional to
\(2P_H\). Therefore,
\[
\mathcal{C}_{\mathrm{sparse}}
=
O\!\left(
(T+2P_L)^2d
\;+\;
Q_H
\left[
T+2(n+2h)^2+(2R+1)^2
\right]d
\right).
\]
The corresponding sparse score storage is
\[
\mathcal{M}_{\mathrm{sparse}}
=
O\!\left(
(T+2P_L)^2
\;+\;
Q_H
\left[
T+2(n+2h)^2+(2R+1)^2
\right]
\right)
\]
for an implementation that stores low/text scores densely and high rows as
CSR. A fused kernel can reduce the constant factor and avoid materializing
the full score tensor.

\subsection{Scaling with resolution}

With \(P_H=n^2P_L\), dense high-resolution visual interactions grow as
\(\Theta(n^4P_L^2d)\). The sparse high-resolution term retains the explicit
support factor
\[
O\!\left(
n^2P_L
\left[(n+2h)^2+(2R+1)^2+T\right]d
\right).
\]
For each fixed 2K/4K run, this is a finite-support edge count. Under the
synchronized rule \(h=Rn\), the support itself scales with \(n\); therefore
the sparse term should not be interpreted as resolution-independent linear
scaling. The parent-cell permutation does not change the edge count; it
improves memory locality and exposes regular CSR rows.

\subsection{End-to-end denoising cost}

At fixed network depth and head count, let the joint-update cost include
attention, projections, and feed-forward computation:
\[
\mathcal{C}_{\mathrm{step}}
=
\Theta\!\left(
\mathcal{C}_{\mathrm{sparse}}+
\mathcal{C}_{\mathrm{proj}}+
\mathcal{C}_{\mathrm{FFN}}
\right).
\]
Let \(\mathcal{C}_{\mathrm{shared}}\) collect common non-denoising work,
including text and source encoding and final VAE decoding. Let
\(\mathcal{C}_{L\text{-only}}\) denote one native 1K update,
\(\mathcal{C}_{\mathrm{switch}}\) the additional 1K velocity prediction at
step \(K\), and \(\mathcal{C}_{\mathrm{lift}}\) the RGB/VAE transition.
The costs are then
\[
\mathcal{C}_{\mathrm{direct}}
=
N\,\mathcal{C}_{\mathrm{step}}+
\mathcal{C}_{\mathrm{shared}},
\]
\[
\begin{aligned}
\mathcal{C}_{\mathrm{progressive}}
={}&K\,\mathcal{C}_{L\text{-only}}+
(N-K)\,\mathcal{C}_{\mathrm{step}}\\
&+\mathcal{C}_{\mathrm{shared}}+
\mathcal{C}_{\mathrm{switch}}+
\mathcal{C}_{\mathrm{lift}}.
\end{aligned}
\]
Thus \(K\) and \(N-K\) count Euler updates, not all model evaluations:
forming the clean estimate requires an additional prediction at the switch.
The lift is paid once, and its overhead must be included when assessing the
benefit of replacing high-resolution updates.

\section{Limitations and Future Work}
\label{app:limitations}

\subsection{Current limitations}

VINS-4KEval lacks pixel-level edit annotations, so evaluation masks are
automatic estimates. GroundingDINO may select a whole object for a part-level
request, and source-only addition regions are conservative when no concrete
anchor is specified. Auditing these masks does not replace human semantic
judgment.

The topology relies on a capable 1K semantic anchor and local target/reference
support with bounded \(\LN\rightarrow\HN\) guidance. Highly nonlocal edits,
large geometric rearrangements, or instructions depending on fine target-scale
text may not fit this division of labor. Boundary-dependent row lengths can
also reduce kernel efficiency when CSR buckets pack poorly.

Progressive inference introduces RGB-transition states that differ from the
native joint trajectory. Although training-free, the path is not explicitly
represented during training. Direct latent interpolation is evaluated only
as a lifting ablation.

\subsection{Future work: training-aware progressive inference}

A natural extension is to include progressive transitions during training.
At a sampled \(K\), a 1K prefix would produce the clean estimate and lifting
would initialize the high-resolution state. Training on a mixture of direct
and post-transition states could reduce the progressive train--test gap
while retaining direct-path quality.

Further directions include source-only mask uncertainty, resolution-adaptive
CSR bucketing, fused neighborhood kernels preserving the single-softmax
union, and human evaluation of part-level edits and small text.

\clearpage
\section{Additional Visualizations}
\label{app:visuals}

\subsection{2K Resolution}
\begin{figure}[h]
\centering
\includegraphics[width=\columnwidth]{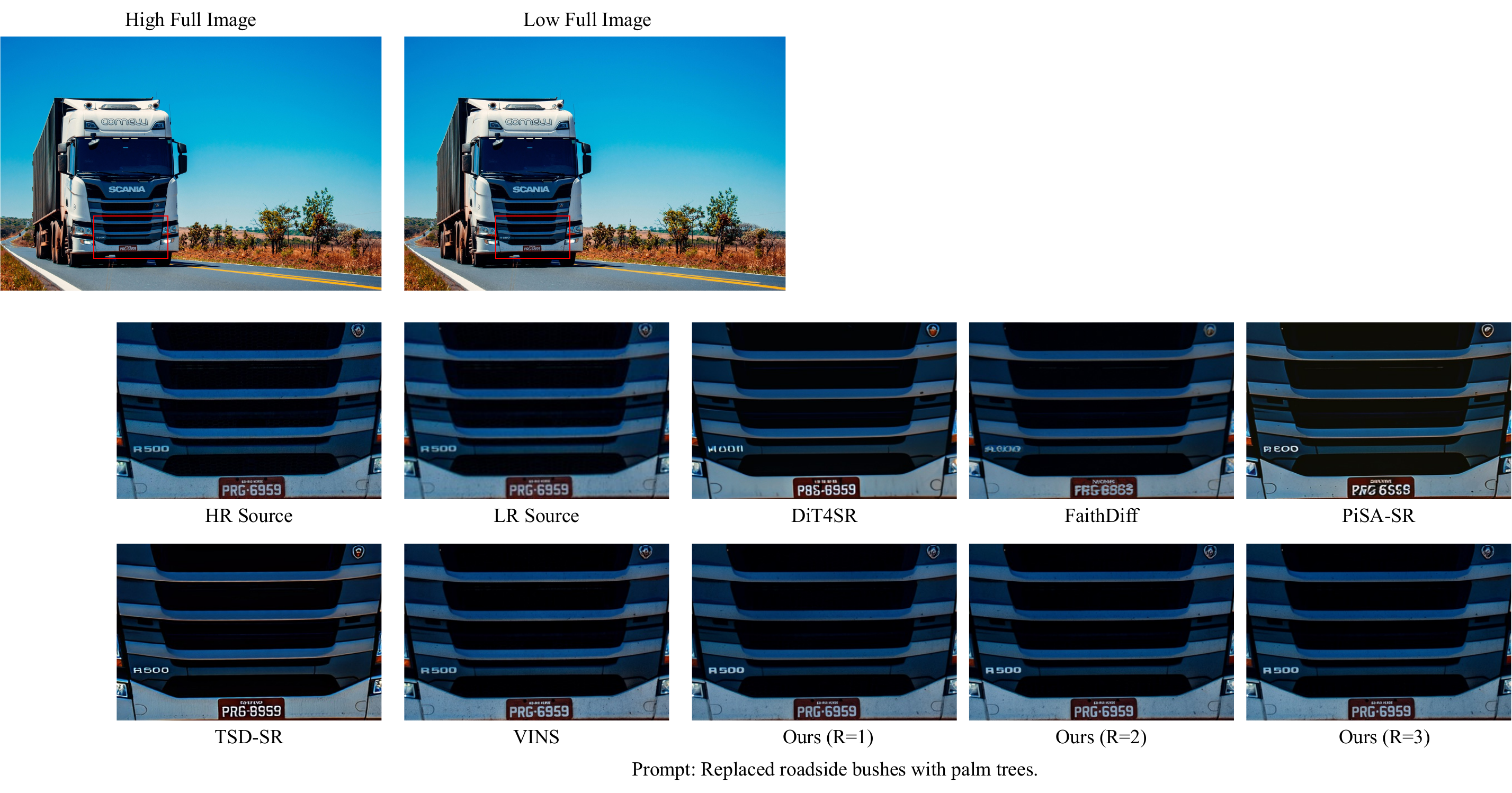}
\caption{Additional 2K comparisons on VINS-4KEval under the shared
GroundingDINO+SAM2 masks. Each group shows the source, instruction, and
method outputs. Local structure, small text, and thin boundaries reveal
differences between 1K-upsampled and native high-resolution editing.}
\label{fig:supp_2k}
\end{figure}

\clearpage
\subsection{4K Resolution}
\begin{figure}[h]
\centering
\includegraphics[width=\columnwidth]{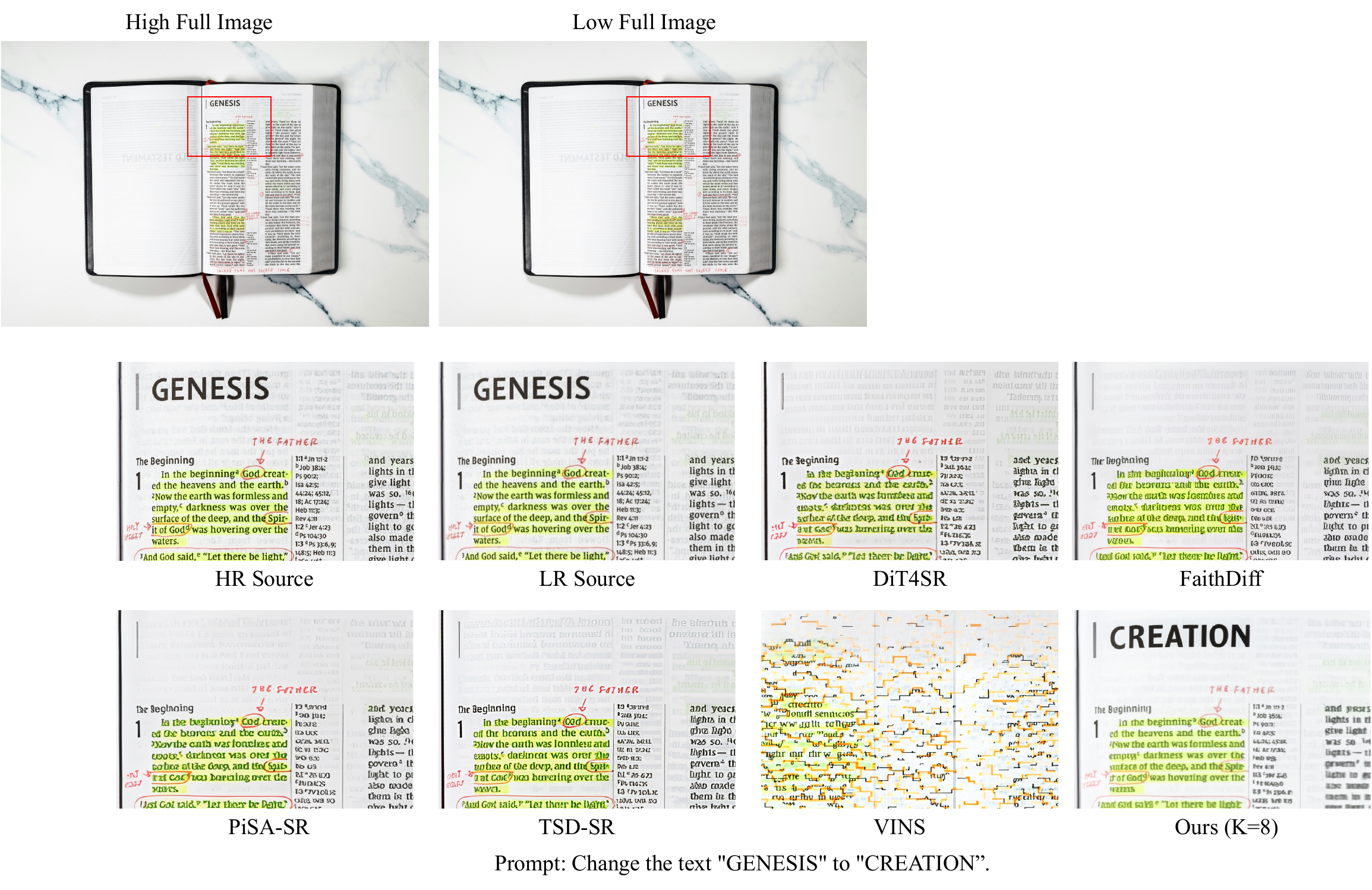}
\caption{Additional 4K comparisons on VINS-4KEval under the same shared-mask
protocol, contrasting native target-resolution detail with 1K editing
followed by super-resolution.}
\label{fig:supp_4k}
\end{figure}


\end{document}